\documentclass{bioinfo}
\usepackage{bm}
\usepackage{graphicx}
\usepackage{xcolor}
\usepackage{appendix}
\usepackage{multirow}
\usepackage{lipsum}
\usepackage{multicol, blindtext}
\usepackage{soul}
\usepackage[capitalize]{cleveref}
\usepackage{xr}
\usepackage{amsmath}
\usepackage{float}
\usepackage[pscoord]{eso-pic}

\copyrightyear{2021} \pubyear{2021}

\access{Advance Access Publication Date: Day Month Year}
\appnotes{Preprint}

\makeatletter
\newcommand*{\addFileDependency}[1]{
	\typeout{(#1)}
	\@addtofilelist{#1}
	\IfFileExists{#1}{}{\typeout{No file #1.}}
}
\makeatother

\newcommand*{\myexternaldocument}[1]{%
	\externaldocument{#1}%
	\addFileDependency{#1.tex}%
	\addFileDependency{#1.aux}%
}

\newcommand{\placetextbox}[3]{
	\setbox0=\hbox{#3}
	\AddToShipoutPictureFG*{
		\put(\LenToUnit{#1\paperwidth},\LenToUnit{#2\paperheight}){\vtop{{\null}\makebox[0pt][c]{#3}}}%
	}%
}%

\myexternaldocument{Pal.29.sup}

\begin{document}
\placetextbox{0.35}{1}{\textsf{This article has been accepted for publication in \textit{Bioinformatics} Published by Oxford University Press.}}%
\firstpage{1}

\subtitle{Subject Section}

\title[integrated REFINED]{Investigation of REFINED CNN ensemble learning for anti-cancer drug sensitivity prediction}
\author[Bazgir \textit{et~al}.]{Omid Bazgir\,$^{\text{\sfb 1}}$, Souparno Ghosh\,$^{\text{\sfb 2}}$ and Ranadip Pal\,$^{\text{\sfb 1,}*}$}
\address{$^{\text{\sf 1}}$Department of Electrical and Computer Engineering, Texas Tech University, Lubbock, Texas, 79409, USA \\
$^{\text{\sf 2}}$Department of Mathematics and Statistics, University of Nebraska-Lincoln, Lincoln, Nebraska, 68583,
USA.}

\corresp{$^\ast$To whom correspondence should be addressed.}

\history{Received on XXXXX; revised on XXXXX; accepted on XXXXX}

\editor{Associate Editor: XXXXXXX}

\abstract{\textbf{Motivation:} Anti-cancer drug sensitivity prediction using deep learning models for individual cell line is a significant challenge in personalized medicine. Recently developed REFINED (REpresentation of Features as Images with NEighborhood Dependencies) CNN (Convolutional Neural Network) based models have shown promising results in improving drug sensitivity prediction. The primary idea behind REFINED-CNN is representing high dimensional vectors as compact images with spatial correlations that can benefit from CNN architectures. However, the mapping from a high dimensional vector to a compact 2D image depends on the a-priori choice of the distance metric and projection scheme with limited empirical procedures guiding these choices.\\
\textbf{Results:} In this article, we consider an ensemble of REFINED-CNN built under different choices of distance metrics and/or projection schemes that can improve upon a single projection based REFINED-CNN model. Results, illustrated using NCI60 and NCI-ALMANAC databases, demonstrate that the ensemble approaches can provide significant improvement in prediction performance as compared to individual models. We also develop the theoretical framework for  combining different distance metrics to arrive at a single 2D mapping. Results demonstrated that distance-averaged REFINED-CNN produced comparable performance as obtained from stacking REFINED-CNN ensemble but with significantly lower computational cost.\\
\textbf{Availability:} The source code and scripts used in the paper have been deposited in GitHub (\href
{Github}{https://github.com/omidbazgirTTU/IntegratedREFINED}).\\
\textbf{Contact:} \href{ranadip.pal@ttu.edu}{ranadip.pal@ttu.edu}\\
\textbf{Supplementary information:} Supplementary data will be available at \textit{Bioinformatics}
online.}

\maketitle

\section{Introduction}

A primary objective of precision medicine for cancer is the selection of an anti-cancer drug or a drug combination that is most effective for the individual patient \citep{garnett2012systematic}. A multitude of methods have been proposed to address the issue of anti-cancer drug sensitivity prediction using high-dimensional genomics or chemical drug descriptors data, but there exists room for achieving significant improvement \citep{chiu2019deep,barretina2012cancer,costello2014community,wan2014ensemble,romm2020artificial}. To offer enhanced predictive performance, numerous deep learning based models have been introduced recently \citep{chiu2019deep,xia2018predicting,chang2018cancer,liu2019improving,keshavarzi2019deepmalaria,yu2019architectures}, that are primarily either deep neural network (DNN) or 1D convolutional neural network (CNN) based approaches. These methods take the input data as a 1-D vector \citep{mostavi2020cancersiamese}, whereas the 2D CNN based method reshape the 1-D vector into a 2D matrix, using some form of lexicographic ordering, which does not preserve the embedded pattern of the data \citep{mostavi2020convolutional}.

We developed the REFINED (REpresentation of Features as Images with NEighborhood Dependencies) \citep{bazgir2020representation} procedure as a general unsupervised isometric mapping to convert high-dimensional vectors into images for training CNN models. We considered a collection of chemical descriptors associated with a drug (or the set of gene expressions associated with a cell line) as a $d-$ dimensional vector of features predicting the efficacy of the drug on a cell line. Thus for $n$ independent drugs (or cell lines), it is a standard univariate high dimensional regression problem. The novelty of our REFINED projection, however, is to represent the foregoing $p-$ dimensional feature vectors (chemical decriptors or gene expressions) as compact images where locally-adjusted Bayesian Multidimensional Scaling, (MDS) solution is used to infer the location of each coordinate of the original high dimensional vector on a bounded subspace of $\mathbb{R}^2$. The dependence among the coordinates of the high dimensional vector induces spatial association in 2D images that is then exploited by the CNN based architecture of the predictive model. We note that REFINED is a general framework that can be applied to any prediction problem involving scalar responses and high dimensional correlated regressors.   

For illustrative purpose, we demonstrated that REFINED-CNN model provided better predictive performance as compared to DNN or 2D random projection based CNN models in publicly available pharmacogenomics data- the NCI60 and GDSC datasets. In the NCI60, we used the chemical descriptors of each drug as input features.   For GDSC dataset, both gene expressions and drug descriptors were used input features \citep{bazgir2020representation}. 

However, in the original form of REFINED, we need to choose a distance metric a-priori to define the ``observed'' distances among the coordinates of the original high dimensional vector and, based on that choice, choose an appropriate projection scheme. For instance, if Euclidean (geodesic) distance is chosen to measure the distances in ambient dimension, MDS (Isomap) is usually chosen to initialize the dimension reduction process. In Figure \ref{DistDiff} (a), we show the distribution of Euclidean distances among chemical descriptors of drugs in ambient dimension and distribution of distances in 2D under various choices of projection schemes for NCI-60 dataset. Observe that, distribution of projected distances obtained under local non-linear dimension reduction approach (LE and LLE) are very different when Euclidean distance is chosen to measure the distance in ambient dimension. If a natural distance metric is not available for the problem at hand, then, ideally, we need to obtain REFINED projections for different distance measures (local versus global, Euclidean versus Geodesic etc.);  obtain the predictions for each candidate distance measure, and choose the one that produces the best cross-validated prediction performance.  Even if an a-priori dissimilarity measure among the coordinates are supplied, we need to choose an appropriate projection scheme (MDS, Isomap etc.) to begin the process of REFINED projection and choose the best initial projection scheme via cross-validation. Evidently, the predictive CNN needs to be fitted for each candidate distance metric/ initial projection schemes, resulting in high computational cost.

Since limited guidelines are available to identify an appropriate choice of distance metric/initial projection schemes in an unsupervised setting,  multiple REFINED-CNNs have to be fitted regardless, resulting in the  availability of an ensemble of REFINED-CNNs. Therefore, a model averaging could be performed which can improve upon the best single REFINED-CNN prediction. The goal of this study is to investigate the performance of such ensemble learners. We illustrate the advantages of three different ensemble methods: (a) model stacking, (b) image stacking, and (c) integrated-REFINED (iREFINED) over the foregoing best single REFINED-CNN predictions. Our key contribution here is the theoretical and methodological development of iREFINED-CNN that produces predictive performance comparable to REFINED-CNN model stacking, but at a considerably lower computation cost. We apply this methodology on NCI60 and NCI-ALMANAC datasets to compare the performance of iREFINED-CNN with several competing methods. Figure \ref{fig:3D_REFINED} illustrates the framework utilized to train each deep CNN model. 

\section{Methodology}

In its original form, REFINED is an unsupervised technique that projects from $\mathbb{R}^d$ to a compact subspace of $\mathbb{R}^2,\;d>>2$. These images are then passed on to a CNN to obtain supervised prediction. Thus, using REFINED images to train a CNN (REFINED-CNN) is, broadly, a 2-step process. This offers an opportunity to deploy ensemble learning in various different ways. In this article, we investigate three such ensembling approaches. We begin with briefly describing the process of creating REFINED images (for more details we direct the audience to \citep{bazgir2020representation}). Next, we describe three ensemble learning approaches we used in this study. Finally, for the sake of completeness, we define the CNN architectures and Bayesian optimization framework that we utilized to select the CNNs' hyper-parameters.

\subsection{REFINED CNN}
\label{REFINED}

REFINED maps high dimensional vectors to mathematically justifiable images for training CNN models. It first uses a user-specified distance metric to obtains the initial pairwise distance matrix for the features in their original space. Then uses Bayesian multidimensional scaling (BMDS) to project the features in 2D that approximately preserve pairwise feature distances in the original space. The resulting initial feature map is then subjected to hill-climbing algorithm with the constraint that each pixel can contain at most one feature. The hill-climbing algorithm essentially provides local adjustements to arrive at a locally optimal configuration which does not produce more distortion as compared to the automorphic solution that BMDS produces. The REFINED algorithm, therefore, uses all the samples to arrive at a set of coordinates that are used to map the features into the target 2D space. Once these locations are fixed, the value of each feature, associated with a particular sample, provide the intensity at the pixel reserved for that feature. For each sample, the algorithm  thus produces unique REFINED image associated with the feature vector for that sample. 

By using different initial distance metric to estimate feature dissimilarity, or choosing different projection schemes to initialize REFINED procedure, different REFINED images could be obtained and consequently the REFINED-CNN's predictive performance vary across the foregoing choices. As shown in \citep{bazgir2020representation}, REFINED CNN initialized with MDS provides better prediction error as compared to Isomap \citep{tenenbaum2000global}, Locally linear embedding (LLE) \citep{roweis2000nonlinear}, and Laplacian eigenmaps (LE) \citep{belkin2003laplacian} on the NCI60 dataset. Therefore, in absence of a natural measure to identify feature dissimilarities and project them to target 2D space, REFINED-CNN needs to be trained for different choices of distance metrics and initial projection schemes.


\subsection{Model stacking}

An immediate consequence of having REFINED-CNN being trained on different choices of distance metrics and initial projection schemes is that we have at our disposal several outputs from the  CNN predictive model each associated with a different choice we made a-priori. Clearly, a linear combination of these predictions, with linear weights estimated from a separate validation set,  produces the REFINED-CNN model stacking. More precisely, let $\Tilde{y}_a$ be the prediction of a REFINED-CNN associated with the choice of a distance metric (or projection scheme) $a=1,2,...A$. Then, the final prediction REFINED-CNN model stacking $Y_f$ is given by the linear regression equation 
\begin{equation}
Y_{f} = \sum_{a=1}^A \gamma_a\Tilde{y}_a + b+\epsilon
\label{modelstacking}
\end{equation}{}
where $\gamma_a$ is the linear weight associated with the choice $a$, $b$ is the intercept term and $\epsilon$ is the error. MLE for regression coefficient could be estimated if the $\epsilon$ is non-Gaussian, but following \citep{costello2014community,wan2014ensemble} we simply use the least square solutions for the regression coeffcients. \citep{kondratyuk2020ensembling} recently showed that, in the context of CNN predictions, ensemble of models usually provide better performance than a single candidate model. We therefore use the model stacking approach to benchmark the performance of other candidate models.

\subsection{Image stacking}

\begin{figure*}[]
	\centering
	\includegraphics[width=\textwidth]{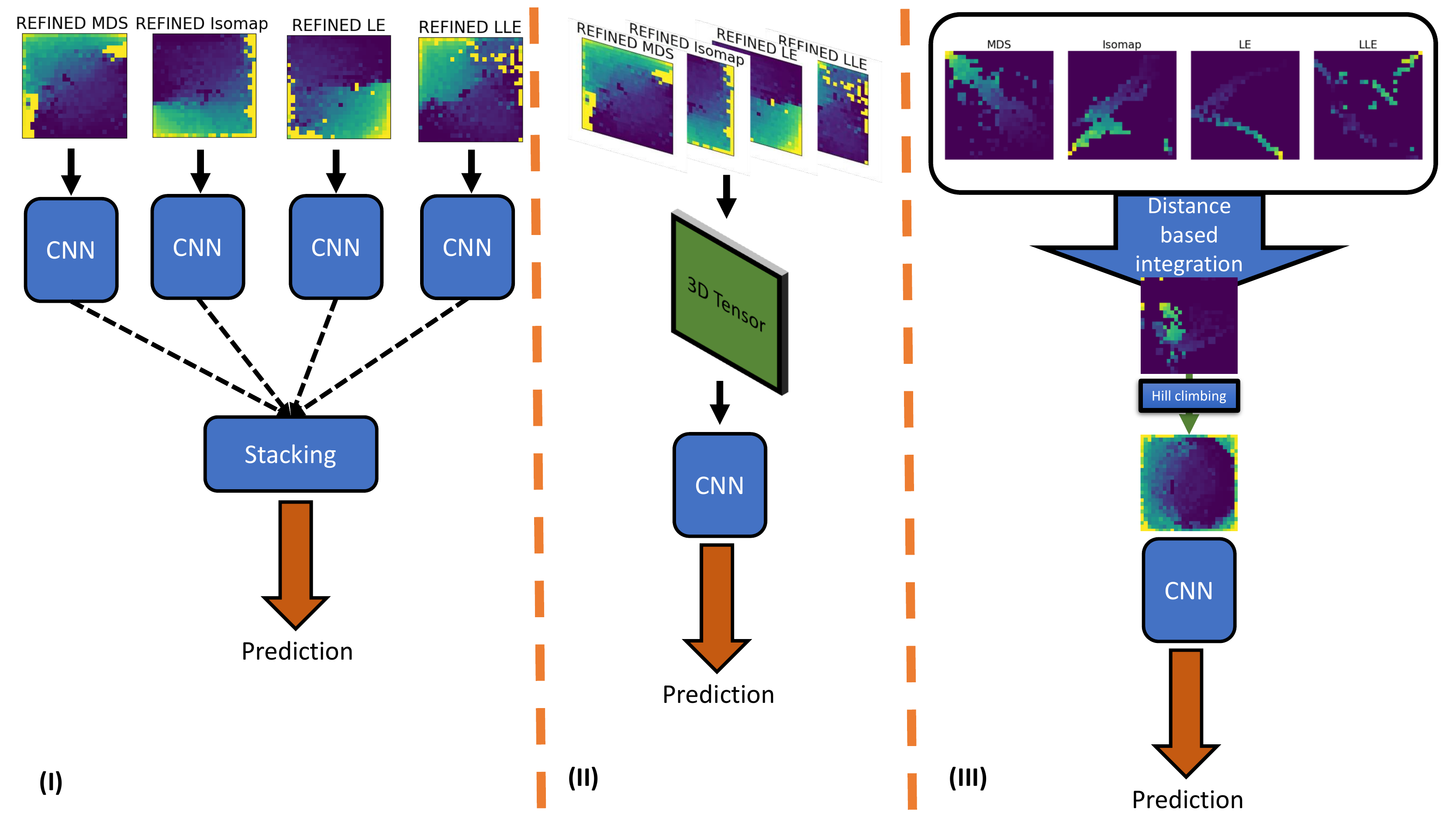}
	\caption{Illustration of three ensemble learning approaches in this study. I) is stacking four different REFINED CNN models to achieve the ultimate prediction. II) is REFINED-CNN image stacking model that stack images in the z-direction prior CNN modeling and III) is integrated REFINED CNN model that integrates all the created REFINED images into one image and then trains a CNN model.}
	\label{fig:3D_REFINED}
\end{figure*}

Evidently, in model stacking approach (\ref{modelstacking}), for each choice $a$, producing the REFINED images $I_a$, separate CNNs need to be trained. Since computational cost associated with CNN training is considerably more than producing $I_a$, one immediate avenue to reduce computation cost is to concatenate the REFINED images $\{I_1,I_2,\cdots,I_A\}$  to produce a 3D tensor for each sample. This 3D tensor can be passed on to the CNN architecture to train a single CNN model using all the images produced by candidate choices. The resulting 3D convolution blocks essentially learns to extract features from the tensors via the back propagation process \citep{ji20123d,maturana2015voxnet}. This approach gets rid of the linearity assumption in (\ref{modelstacking}) and model averaging is done implicity. Additionally, it requires training of a single CNN thereby reducing the computation cost significantly. In the context of this study, a graphical representation of anti-cancer drug sensitivity prediction with 3D convolution blocks is shown in Figure \ref{fig:3D_REFINED}. 
\begin{figure}[]
	\centering
	\includegraphics[width=0.5\textwidth]{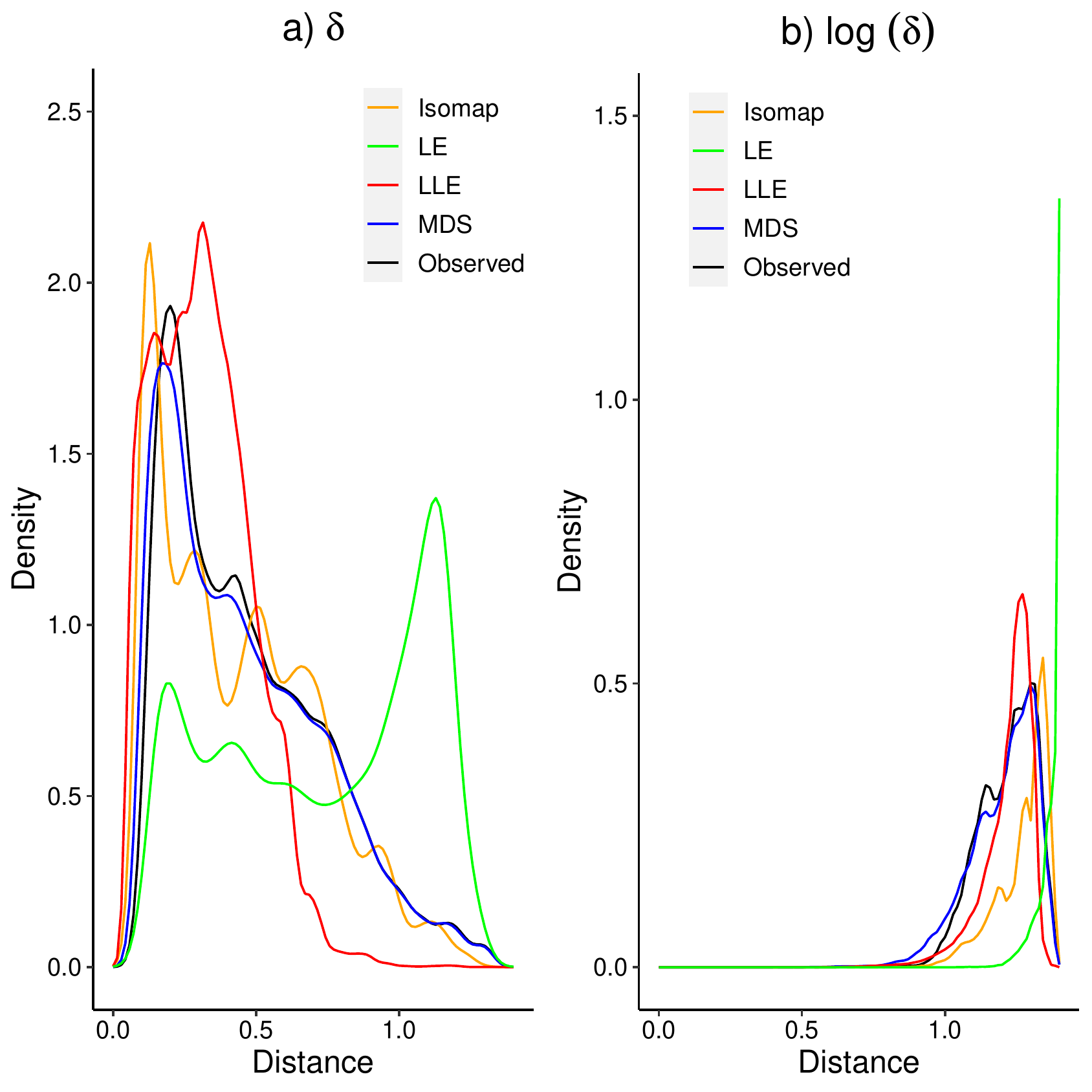}
	\caption{Density of distances. Kernel density estimate of between features' observed (Euclidean) distance versus distances of projection in 2D space by the 4 DR techniques in regular and log scale.}
	\label{DistDiff}
\end{figure}

Although this technique offers computational benefits, it still requires generation of ''$A$'' REFINED images. More importantly, since each $I_a$ is created independently for each choice, and because locations are not uniquely identifiable in BMDS solution, there is no guarantee that a particular feature will occupy the same coordinate in each $I_a, a=1,2,...,A$. Consequently, when the images are concatenated, a particular coordinate often does not correspond to an unique feature across $I_a$, thereby severely affecting CNN's ability to extract features from the input tensors. Lack of coordinate-specific association of pixel intensities across $I_a$ also potentially impacts the predictive performance of the CNN. 

To partially address the lack of uniqueness in feature locations across REFINED images, we stack the feature maps extracted by the convolution layers instead of stacking the raw REFINED images. Towards that end, for each REFINED image, we design the convolution layer under different choices of the number and size of kernels. By allowing the kernel sizes to vary across REFINED images we can potentially capture the impact of distance metrics defined over different scale, i.e., the global vs local nature of MDS/Isomap and LE/LLE, respectively.   The feature maps extracted by these convolution layers are then concatenated and passed on to the dense layers. The details of the feature map stacking is provided in the \cref{sec:featuremap} of the supplementary information.

Regardless, to fully alleviate this context-specific non-uniqueness problem, we need to enforce the condition that location of each feature remains same for all input REFINED images. The integrated REFINED methodology arises when this condition is enforced to infer the location of each feature.

\subsection{Integrated REFINED}
Consider the predictor matrix $\bm{X}=\{x_{ij}\}, i=1,2,..., n; \; j=1,2,...,p$ with $x_{ij}$ being the value of the $j$th feature for the $i$th sample. The goal of REFINED was  to obtain the location of the features in a compact subset of $\mathbb{R}^2$, more specifically in $[0,1]^2$. In the following formulation, we assume each choice of initial distance metric is uniquely associated with a projection scheme leading to a total of $A$ choice of distance metric-projection schemes pairs. Let $d_{jk,a}$ be the observed distance between the $j$th and the $k$th feature obtained using the distance metric $a$ and $\delta_{jk}$ be the unknown Euclidean distance between these two features in unit square. Hence, $\delta_{jk}=\sqrt{\sum_l (s_{j,l}-s_{k,l})^2}$, where $s$ is now 2D coordinate system denoting the unique location of the features $j$ and $k$ in unit square obtained by synthesizing $d_{jk,a}, a=1,2,...,A$. Our goal is to estimate $s_j \in [0,1]^2$ that remains invariant for all candidate distance metric.

Under the assumption of truncated normal distribution of $d_{jk,a}$ \citep{oh2001bayesian}, the data model associated with the distance metric $a$ is given by $d_{jk,a}\sim N(\delta_{jk},\sigma_a^2)I(d_{jk,a}>0)$. For the location process, we specify a spatial Homogeneous Poisson Process (HPP) with constant intensity $\lambda=p/[0,1]^2$ which essentially distributes locations of $p$ predictors randomly in an unit square. Since this corresponds to complete spatial randomness, an alternative specification of location process is given by $\bm{s}=\{s_1,s_2,...,s_p\}\sim Uniform([0,1]^2)$ \citep{chandler2013spatially}. The advantage of this HPP specification for the location process is outlined in \citep{bazgir2020representation}.

Let $\bm{d}_a=[d_{jk,a}],\; j,k=1,2,...,p$ be the collection of $m$=$p\choose 2$ distances obtained under the metric $a$ and  $\bm{d}=[\bm{d}_1,\bm{d}_2,...,\bm{d}_A]$ be the total number of distances in the dataset. Let  $\bm{\delta}$ be the collection of Euclidean distances in the unit square that needs to be inferred after imposing the invariance of $\bm{s}$. Then under the assumption of conditional independence, the full data model is then given by
\begin{equation}
\resizebox{.9\hsize}{!}{$f(\bm{d}|\bm{s},\sigma^2) (\Pi\sigma^{2}_{a})^{-\frac{m}{2}} e^{-\frac{1}{2}\sum_{j>k}(\sum_{a}(\frac{d_{jk,a}-\delta_{jk}}{\sigma_{a}})^{2})}.e^{-\sum_{a}\sum_{j>k}\log\Phi(\frac{\delta_{jk}}{\sigma_{a}})}$}
\label{likelihood}
\end{equation}
where $\Phi(.)$ is the usual standard normal cdf.
At the process level, we have 
\begin{equation}
\bm{s}|p  \sim Uniform([0,1]^2)
\end{equation}
Finally, we impose the same prior for $\bm{\sigma}^2= [\sigma_1^2,\sigma_2^2,...,\sigma_A^2] \stackrel{iid}{\sim}  \text{Inverse Gamma}(\alpha,\beta)$ with  $a>2,\;\; b>0$. 

Under this specification, the full conditional of posterior of $\bm{s}$ is given by 
\begin{equation}
\pi(\bm{s}|\bm{d},\bm{\sigma}^{2}) \propto e^{\frac{-1}{2}\sum_{j>k}V(\delta_{jk}-\frac{\overline{d}_{jk0}}{V})^{2}}.e^{-\sum_{a}\sum_{j>k}\log\Phi(\frac{\delta_{jk}}{\sigma_{a}})}   
\label{posterioram}
\end{equation}
where $V=\sum^{A}_{a=1}\frac{1}{\sigma^{2}_{a}} \& \overline{d}_{jk0} = \sum^{A}_{a=1}\frac{d_{jk,a}}{\sigma^{2}_{a}}$. They key observation is that, the location parameter of the conditional posterior of $\bm{s}$ is the weighted average of the observed distances obtained from each distance metric under consideration, with the weights being a function of  the precision associated with the distribution of observed distances. The details of the derivation of (\ref{posterioram}) is relegated to the \cref{A1} of the supplementary information.

If, on the other hand, we posit a log-normal distribution for $d_{jk,a}$ \citep{bakker2013bayesian}, the data model associated with the distance metric $a$ is given by $\log(d_{jk,a})\sim N(\log(\delta_{jk}),\sigma_a^2)$. Retaining the HPP specification of location process and independent Inverse Gamma priors for $\bm{\sigma}^2$, the posterior conditional of $\bm{s}$ is given by
\begin{equation}
\pi(\bm{s}|\bm{d},\bm{\sigma}^{2}) \propto   e^{-\frac{1}{2}\sum_{j>k}V(\delta^{*}_{jk}-\frac{\overline{d}^{*}_{jk0}}{V})^{2}}
\label{posteriorgm}
\end{equation}
where $\delta^{*}_{jk}=\log(\delta)_{jk}$,  and $\overline{d}^{*}_{jk0}=\sum^{A}_{a=1}\frac{\log(d)_{jk,a}}{\sigma^{2}_{a}}$. Further simplification of the location parameter in (\ref{posteriorgm}) yields $\frac{\overline{d}^{*}_{jk0}}{V} = \frac{\sum W_{a} \log d_{ija}}{\sum W_{a}}$, where $W_a=\frac{1}{\sigma^2_a}$. Clearly, the location parameter  is the weighted geometric mean of $(d_{jk1},...,d_{jkA})$ with the weight being a function of precision associated with the distributional specification of $d_{.,a}$. Detailed derivations of (\ref{posteriorgm}) is offered in the \cref{A2} of the supplementary information. 

Observe that,  (\ref{posterioram}) and (\ref{posteriorgm}) imply that one of the  ways to fix the coordinate associated with each feature across $I_a$, is to enforce a common $\delta(.)$ for data model associated with each distance metric. The methodoligical benefits of iREFINED are twofold: (a) feature-specific coordinates $\bm{s}$ can be estimated using standard BMDS solutions without explicitly specifying a composite dissimilarity measure (linear combination of initial distances either in original scale or in log scale) at the outset, and (b) if a linear combination of the candidate distance metrics is utilized to obtain the initial dissimilarity measure at the outset, Bayesian non-metric MDS can be performed with a suitable choice of a monotonic nonlinear function $g(.)$ that connects the observed dissimilarities with $\delta$ in the following way $d_{.}\sim N(g(\delta(.)),\sigma^2)I(d(.)>0)$ \citep{oh2001bayesian}. The fact that the composite dissimilarity measure may not be proper metric is accommodated by an explicit non-metric BMDS formulation. The computational benefit of iREFINED-CNN is obvious, it requires a single REFINED projection obtained from the estimates of $\bm{s}$ given by  (\ref{posterioram}) or (\ref{posteriorgm}) which is subjected to the foregoing hill-climbing algorithm to arrive at single REFINED image which is then passed on to a single CNN. Consequently, regardless of the number of choice of initial  distance metrics (and the associated initial projection schemes), iREFINED-CNN only requires a single full-blown training operation.

\section{Application}

We apply the methodologies developed in the previous section on two publicly avaiable datasets: (a) \textit{NCI60} dataset consists of drug responses observed after application of more than 52,000 unique compounds on 60 human cancer cell lines \citep{shoemaker2006nci60},  (b) \textit{NCI-ALMANAC} dataset consisting over 5,000 pairs of more than 100 drug responses on 60 human cancer cell lines \citep{holbeck2017national}. In both scenarios, we use the chemical descriptors of drugs as features to  predict cell-line specific drug responses. Below we offer brief description of each dataset, outline individual REFINED projection schemes to formulate the iREFINED procedure and describe the CNN architecture.

\subsection{Data description}

{\bf NCI60:} The US National Cancer Institute (NCI) screened more than 52,000 unique drugs on around  60  human  cancer  cell  lines. The drug responses  are  reported  as  average growth inhibition of 50 \% (GI50) across the entire NCI cell panel \citep{GERSON2018849} \citep{shoemaker2006nci60} . All the chemicals have an associated unique NSC identifier number. We used the NSC identifiers to obtain the chemical descriptors associated with each drug. This information was supplied to PaDEL software \citep{yap2011padel} to extract relevant features for each one of the foregoing chemicals. Chemicals with more than 10 \% of their descriptor values being zero or missing were discarded. To ensure availability of enough data points for training deep learning models, we selected 17 cell lines with more than 10,000 drugs tested on them. Each drug was described with 672 features. To incorporate the logarithmic nature of dose administration protocol, we calculated the normalized negative-log concentration of GI50s (NORMLOGGI50). The drug response distribution for three illustrative cell lines are shown in \cref{NCI60_Distribution} of the supplementary information.

\begin{figure}[]
	\centering
	\includegraphics[width=0.5\textwidth]{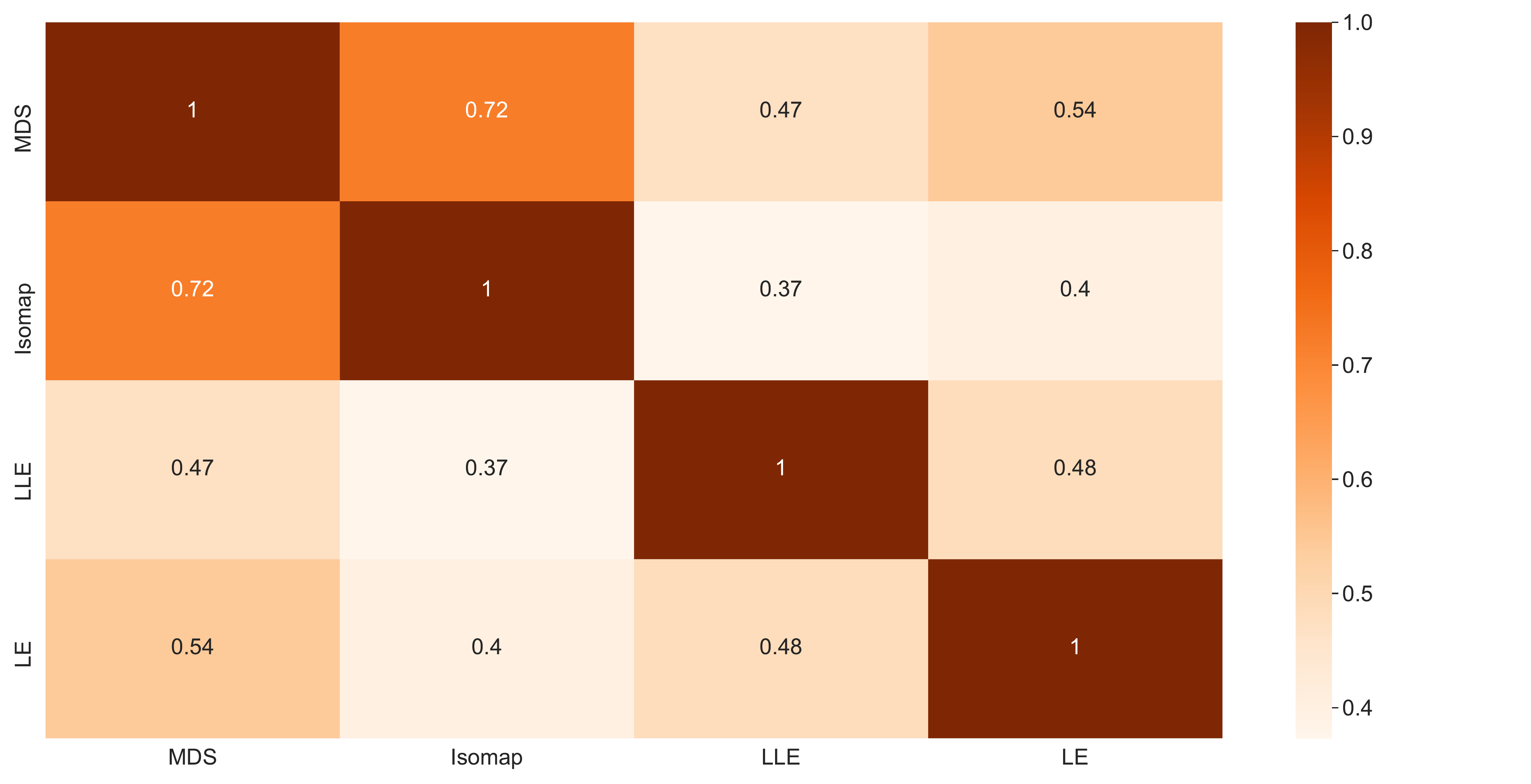}
	\caption{Correlation between distances. Kendall's $\tau$  among the distances estimated in 2D by each DR technique and their geometric and arithmetic means.}
	\label{DisHeat}
\end{figure}

We considered four distance metric and associated projection schemes- MDS, Isomap, LE, and LLE- to initialize the REFINED process.
To investigate if these four techniques produce similar ordering of the pairwise distances between features, we calculated the following Kendall's rank correlation coefficients  $\tau(R({d}_{jk,a}),R({d}_{jk,a'})),\;\; \forall j,k=1,2,...,p, \text {and } a\neq a'=1,2,3,4$ where $R({d}_{jk,a})$ is the rank of the distance between features $j$ and $k$ obtained from the projection technique $a$. 
Figure \ref{DisHeat} shows the heat map of the foregoing rank correlations. Evidently, there is a strong agreement between MDS and Isomap. But only moderate level of association between the global techniques and local techniques. However, the distribution of observed Euclidean distance in log-scale in Figure \ref{DistDiff} (b) shows better agreement with the logarithm of projected distances, across both local and global dimension reduction schemes indicating the viability of
log-normal specification of the data model in the foregoing iREFINED technique. Furthermore, \cref{tab:KL} shows the Kullback-Liebler divergence between observed Euclidean distance and projected distances in both original scale and log-scale. Observe that, on an average, the KL divergence in log-scale is smaller than that in the original scale, indicating that the log-normal specification offers some protection against misspecification of the initial projection scheme.

\begin{figure}[]
	\centering
	\includegraphics[width=0.5\textwidth]{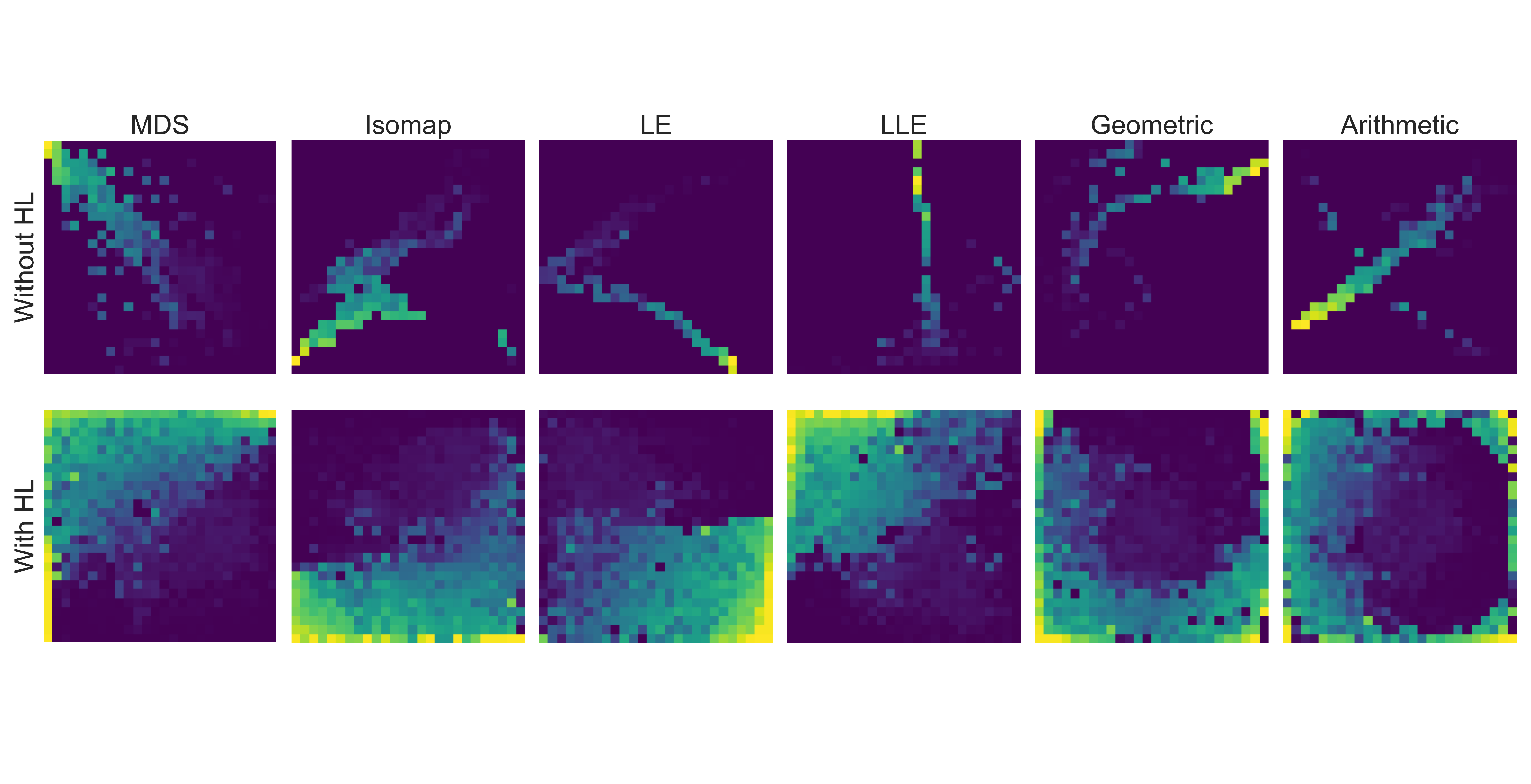}
	\caption{Different REFINED images. REFINED images created using 4 DR technique including MDS, Isomap, LLE, LE, and arithmetic and geometric average of them as initialization step at the first row before applying the hill climbing. The second Row represents the REFINED images after applying the hill climbing algorithm on each initialization step.}
	\label{fig:Geometric}
\end{figure}

While the first four panels of Figure \ref{fig:Geometric} show the REFINED images of drug chemical descriptors created under various initializations for cell line SNB\_78, last two panels show the corresponding iREFINED images under log-normal and truncated normal specifications, respectively.    

{\bf NCI-ALMANAC:} The NCI-ALMANAC is ``A Large Matrix of Anti-Neoplastic Agent Combinations'' dataset \citep{holbeck2017national} provides systematic evaluation of over 5,000 pairs of 104 FDA-approved anticancer drugs were scanned against a panel of 60 human tumor cell lines (from NCI60) to discover those with enhanced growth inhibition or cytotoxicity profiles \citep{yang2020stratification}. Combination activity was reported as a "ComboScore" that quantifies the advantage of combining two drugs \citep{tavakoli2019learning}. Normalized growth percentage of ComboScore distribution for three cell lines selected randomly from NCI-ALMANAC dataset are shown in \cref{NCI_ALM_Distribution} of the supplementary information. For each drug we used the same chemical descriptors obtained for NCI60 dataset using the NSC identifiers.


{\bf CNN architecture:} 
We had two different CNN architectures; one for modeling the NCI60, and another for NCI-ALMANAC dataset. The REFINED CNN used to model NCI60 dataset, contains two convolutional and two fully connected (FC) hidden layers where each followed by a batch normalization (BN) and ReLu activation function layer. Each ReLu activation after the FC layers was followed by a dropout layer to avoid overfitting. 

The REFINED CNN models of NCI-ALMANAC dataset, which predict the ComboScore of two drugs, contain two input as two different drugs in two arms. Each arm contains two convlotional layers followed by a BN and ReLu activation layer. The two arms' output then concatenated and flattened as a 1-D vector as an input of two sequential FC layers, each followed by a BN, ReLu activation function and a dropout layer. 

The hyper-parameters of both these CNN models, i.e., learning rate, decay rate, decay step of the adam optimizer, number of kernels, kernel size, stride size per each convolutional layer, and number of nodes per each fully connected layer, were optimized using Bayesian optimization framework \citep{bergstra2013making,bazgir2020active} which sequentially queries a posterior model for  hyper-parameter $\Theta$ derived from a sequence of surrogate models. 

The hyper parameters of the CNN were optimized, for each dataset, using the training and validation set of only one cell line (HCC-2998). Then the model was trained and tested on each cell line independently. In the test phase, for each cell line, we held out a separate set of drugs in the NCI60 and separate set of drug pairs in the NCI-ALMANAC dataset.

\section{Results}
Several competing models were trained on the foregoing NCI60 and NCI-ALMANAC dataset. Each model was fitted separately on the drug-response data for each cell line. For each cell line, the data was randomly partitioned into training, validation and test sets. Training set consisted of 80\% of the sample, 10\% of the samples were used for validation and the remaining 10\% formed the test set to evaluate the out-of-sample predictive performance of the competing models. To ensure direct comparability, the training, validation and test datasets remained same for all competing models.

A total of 11 models \footnote{A summary description of the baseline models are shown in \cref{tab:ComModels} of the supplementary information.} were considered: (a) Ensemble REFINED-CNN model stacking, (b) Ensemble REFINED CNN-image stacking model, (c) iREFINED-CNN, with both weighted arithmetic mean and weighted geometric mean construction, (d) individual REFINED CNN with MDS, Isomap, LLE, and LE projections, (e) DeepSyenrgy \citep{preuer2018deepsynergy}, (f)  \citep{xia2018predicting} approach, (g) Gradient Boosting Machine \citep{friedman2002stochastic}, (h) Random Forests \citep{ho1995random}, (i) Support Vector Regression \citep{drucker1997support}, (j) Kernelized Bayesian Multitask Learning (KBMTL) \citep{gonen2014drug}, and (k) Elastic Nets \citep{zou2005regularization}. We only applied the DeepSyenrgy \citep{preuer2018deepsynergy} and the \citep{xia2018predicting} approaches on the NCI-ALMANAC dataset, as they are designed for drug combination therapy modeling.  We emphasize that all the competing models were independently used for prediction task.  Although, GBM, SVR, RF, KBMTL could be used as nonlinear/model-free stacking devices to combine the output of different individual REFINED CNNs, we did not pursue that that avenue here.

Several performance measures were used to assess the adequacy of the proposed models and compare their predictive performances. Below we describe the metrics used to evaluate the model performance:
\begin{itemize}
	\item[1.] Normalized root mean square error of prediction (NRMSE): The customary root mean squared error of prediction (RMSPE) of a given model was normalized by the RMSPE with sample mean as the predictor. We use NRMSE to implicitly compare all the models with respect to the baseline intercept-only model. The NRMSE formula is given by:
	
	\begin{equation} \label{NRMSE}
	\text{NRMSE of a model}  = \sqrt{\frac{\sum_{i=1}^{n_p}(y_{i}-\hat{y_{i}})^2}{\sum_{i=1}^{n_p}(y_{i}-\bar{y})^2}}
	\end{equation}
	
	where the$n_p$ is the size of test-set, $y$, $\bar{y}$ , and $\hat{y}$ are the observed drug response, mean of the drug responses obtained from the non-test set, and predicted drug responses obtained from the model under consideration .
	
	\item [2.] Normalized mean absolute error (NMAE): In addition to NRMSE, we use NMAE (\ref{NMAE}) so that model comparison can be performed without being severely impacted by large outliers. 
	\begin{equation} \label{NMAE}
	NMAE = {\frac{\sum_{i=1}^{n_p}\mid y_{i}-\hat{y_{i}}\mid}{\sum_{i=1}^{n_p}\mid y_{i}-\bar{y}\mid}}
	\end{equation}
	For both NRMSE and NMAE, smaller values indicate better predictive performance. 
	
	\item[3.] Pearson correlation coefficient (PCC) between the predicted and target values: PCC quantifies linear association between the predicted and target drug responses. Model with PCC closer to 1 would be preferred. 
	
	\item[4.] Bias reduction: We use the method described in \citep{song2015bias,bazgir2020representation} to compute model bias. A simple linear regression is performed between residuals (ordinate) and predicted values (abcissa) in the test set. The angle ($\theta$) between the best fitted regression line and abscissa is used as a measure for bias. An unbiased model is expected to produce an angle of $0^{\circ}$. Therefore, models with  smaller value of $\theta$ is preferred.

	\item[5.] Model improvement: We introduce a novel measure for model improvement that uses Gap statistics \citep{tibshirani2001estimating} to perform a formal hypothesis test. 
	First, we paired each model with a null model (see \citep{costello2014community} for the construction of null model). Then bootstrap samples were drawn from the drug response values of the test set along with their corresponding predicted values for each model. The null model, using the distribution of drug responses in the training set, is then used to predict drug response sampled from the test set. The process is repeated for 10,000 times and a distribution of NRMSE, NMAE, PCC, and Bias, is made for each model along with the null model. For each candidate model, the bootstrapped distribution of each metric is paired with the corresponding distributions obtained from the null models. 
	
	A model is deemed to provide significant statistical improvement over the null model if each performance metric is stochastically {\it better} than its counterpart obtained from the null model. Therefore, we concatenated the bootstrap replicates of performance metrics under the candidate model and null model and formally tested for the presence of at least two clusters using gap statistics in a completely unsupervised fashion. If gap statistics identified presence of at least 2 clusters, we performed K-means clustering. Ideally, the clustering procedure should be able to distinguish replicates coming from null model and candidate model. Hence, an adequate model will produce little overlap between the clusters associated with the candidate model and those associated with the null model. Additionally, all models were subjected to a robustness analysis \citep{costello2014community}, where we calculated how many times each ensemble REFINED model outperforms other competing models in 10,000 repetition of bootstrap sampling process \citep{bazgir2020representation}
	
	We calculated 95 \% confidence interval for each of the foregoing performance metrics using a pseudo Jackknife-after-Bootstrap confidence interval generation approach \citep{efron1992bootstrap}. Multiple bootstrap sets were drawn from the test samples and then the model performance metrics calculated resulting in a distribution for each metric which was used to calculate the confidence interval for a given cell line for NCI60 and NCI-ALMANAC datasets \citep{bazgir2020representation}.
	
\end{itemize}

\subsection{ Results for NCI60}

First, we report the performance of the nine candidate models, averaged over 17 cell lines, in Table \ref{NCI_SOTA}. Observe that, although we expected that REFINED-CNN model stacking will perform best, it was not uniformly better in terms of all the evaluation metrics. Two variants of iREFINED-CNN produced better performance with respect to NMAE and Bias reduction. The REFINED-CNN image stacking performed uniformly worse as compared to the remaining ensemble REFINED models. One of the reasons for this worse performance could be the inability of image stacking approach to extract appropriate features across the REFINED images. However, all the ensemble variants uniformly outperformed single projection based REFINED models and other popular machine learning models considered here.  
The \cref{tab:NCI60_res_a,tab:NCI60_res_b,NCISOTA_Det} of the supplementary information details the performance of each model with respect to the foregoing metrics for different cell lines.  
The 95 \% confidence interval for all the models per each cell line are provided in \cref{NCI_NRMSE_Conf,NCI_NMAE_Conf,NCI_PCC_Conf,NCI_Bias_Conf} of the supplementary information.

In terms of improvements in prediction, we observe that  REFINED-CNN model stacking decerased NRMSE, NMAE and bias by 7-9\%, 6-9\% and 1-2\%, respectively, as compared to single REFINED model. The former ensemble model also increased the PCC by 6-9\% as compared to the latter. Integrated REFINED decerased NRMSE, NMAE and bias by 6-8\%, 7-12\% and 3-4\%, respectively, and increased PCC by 5-8\% as compared to single REFINED model. However, REFINED-CNN image stacking merely decerased NRMSE, NMAE and bias by 1-3\%, 2-7\% and 0-1\%, respectively, and increased PCC by 1-3\% as compared to single REFINED model, indicating its inability to compete favorably with the previous two ensembling approached 


Turning to  robustness analysis to compare integrated REFINED and REFINED-CNN model stacking models with other single REFINED CNN models, we observe that REFINED-CNN model stacking offers better performance in terms of (a) NRMSE between 73-80\% of the times, (b) NMAE 71-81\% of the times, (c) PCC 68-76\%of the times , and (d) Bias 48-56\% of the times (see \cref{Rob_Stack_NRMSE,Rob_Stack_NMAE,Rob_Stack_PCC,Rob_Stack_Bias}). The integrated REFINED, on the other hand, produced better performance in terms of (a) NRMSE between 70-78\% of the times, (b) NMAE 77-87\% of the times, (c) PCC between 67-75\% of the times, and (d) Bias 55-63\% of the times on average as compared to other single REFINED CNN models (see \cref{Rob_AM_NRMSE,Rob_AM_NMAE,Rob_AM_PCC,Rob_AM_Bias,Rob_GM_NRMSE,Rob_GM_NMAE,Rob_GM_PCC,Rob_GM_Bias}).
The Gap statistics also indicate that the out-of-sample performance metrics produced by REFINED-CNN model stacking and integrated REFINED are, on average, well distinguishable from the null model (see \cref{tab:GapStatNCI60_error1,tab:GapStatNCI60_error2,tab:GapStatNCI60_PCC,tab:GapStatNCI60_Bias} of the supplementary information). Furthermore, higher values of the Gap statistics associated with ensemble models as compared to those associated with single REFINED-CNN models indicate higher degree of separation of the performance metrics clusters associated with ensemble models from the null model as compared to the single REFINED-CNN versions. The NRMSE, NMAE, PCC, and Bias distribution of all the eight models along with the null model are plotted for three randomly chosen cell lines of the NCI60 dataset in \cref{GapDist_NRMSE1,GapDist_NMAE1,GapDist_PCC1,GapDist_Bias1,GapDist_NRMSE2,GapDist_NMAE2,GapDist_PCC2,GapDist_Bias2,GapDist_NRMSE3,GapDist_NMAE3,GapDist_PCC3,GapDist_Bias3} of the supplementary information.


%

In addition to intra-REFINED comparisons, we compare our REFINED-based approaches with state-of-the-art models including: Kernelized Bayesian Multitask Learning (KBMTL) \citep{gonen2014drug}, Gradient Boosting Machine \citep{friedman2002stochastic}, Random Forests \citep{ho1995random}, Support Vector Regressor \citep{drucker1997support}, and Elastic Nets \citep{zou2005regularization}. The average performance of all the models on NCI60 dataset are provided in Table \ref{NCI_SOTA}. Observe that, on average, REFINED-based models significantly outperforms the competing non-REFINED models. The same trend is observed for most cell-lines as well. The detailed results including performance of each model for each cell line is provided in \cref{NCISOTA_Det} of the supplementary information.  
\begin{table}[t]
	\centering
	\caption{NCI60 results. Comparison of performance of proposed approaches, single projection based REFINED (sREFINED), and state-of-the-art methods on NCI60 dataset. The bold values indicate best performance.}
	\label{NCI_SOTA}
	\begin{tabular}{ccccc}
		\hline
		Model          & NRMSE & NMAE  & PCC   & Bias  \\
		\hline
		REFINED-CNN model stacking   & \textbf{0.702} & 0.653 & \textbf{0.710} & 0.489 \\
		iREFINED-CNN-AM    & 0.715 & \textbf{0.630} & 0.706 & 0.461 \\
		iREFINED-CNN-GM    & 0.722 & 0.635 & 0.705 & \textbf{0.446} \\
		REFINED-CNN image stacking & 0.775 & 0.679 & 0.655 & 0.509 \\
		sREFINED with Isomap & 0.787 & 0.716 & 0.644 & 0.509 \\
		sREFINED with LE & 0.788 & 0.720 & 0.644 & 0.504 \\
		sREFINED with LLE & 0.795 & 0.759 & 0.625 & 0.511 \\
		sREFINED with MDS & 0.778 & 0.709 & 0.650 & 0.488 \\
		KBMTL \citep{gonen2014drug}      & 0.856 & 0.768 & 0.547 & 0.733 \\
		XGBoost \citep{friedman2002stochastic}  & 0.842 & 0.806 & 0.513 & 0.781 \\
		SVR  \citep{drucker1997support}  & 0.870 & 0.806 & 0.525 & 0.755 \\
		RF   \citep{ho1995random}        & 0.880 & 0.846 & 0.486 & 0.816 \\
		EN  \citep{zou2005regularization}& 0.976 & 0.942 & 0.287 & 0.968 \\
		\hline
	\end{tabular}%
\end{table}

\subsection{NCI-ALMANAC}

In this section, we compare the performance of the foregoing three ensemble REFINED-CNN approaches with 4 single REFINED-CNN methods utlizing different projection schemes along with 6 non-REFINED predictive methods. Since this dataset offers information about responses for drug combinations, our predictors consist of two set of PaDel chemical descriptors representing two drugs for each cell line. The response consists of the \textit{"ComboScore"} for each drug pair. We used the REFINED approach to generate the images corresponding to the drug descriptors for each drug compound in the NCI-ALMANAC dataset.

Considering pairing 2 drugs with D ($\sim$ more than 100) unique NSCs for each cell line, then the total number of samples for modeling each cell line is $ \binom{D}{2} $ pairs in the dataset, which is close to 5K. For each cell line, we randomly divided the dataset into 80\% training, 10\% validation and 10\% test sets, where each set covariates contains 672 chemical drug descriptors per each drug. 
REFINED-CNN model stacking and integrated REFINED CNN model outperforms all other four single REFINED CNN models whereas REFINED-CNN image stacking under-performs them in average. The REFINED-CNN model stacking and integrated REFINED CNN model achieve improvement over single REFINED CNN models in the range of: 7-10\% and 2-5\% for NRMSE; 8-12\% and 1-5\% for NMAE; 2-3\% and 1-2\% for PCC; 6-12 \% and 1-4\% for Bias. The 95 \% confidence interval for all the models per each cell line are provided in \cref{ALM_NRMSE_Conf,ALM_NMAE_Conf,ALM_PCC_Conf,ALM_Bias_Conf} of the supplementary information.
%

\begin{table}[t]
	\centering
	\caption{NCI-ALMANAC results. Comparison of performance of proposed approaches, single projection based REFINED (sREFINED), and state-of-the-art methods on NCI-ALMANAC dataset. The bold values indicate best performance.}
	\label{ALM_SOTA}
	\begin{tabular}{ccccc}
		\hline
		Model          & NRMSE & NMAE  & PCC   & Bias  \\
		\hline
		REFINED-CNN model stacking   & \textbf{0.420} & \textbf{0.361} & \textbf{0.907} & \textbf{0.168} \\
		iREFINED-CNN-AM    & 0.479 & 0.431 & 0.893 & 0.275 \\
		iREFINED-CNN-GM    & 0.474 & 0.427 & 0.892 & 0.248 \\
		REFINED-CNN image stacking & 0.561 & 0.524 & 0.856 & 0.362 \\
		sREFINED with Isomap & 0.508 & 0.470 & 0.887 & 0.227 \\
		sREFINED with LE & 0.489 & 0.443 & 0.884 & 0.238 \\
		sREFINED with LLE & 0.522 & 0.486 & 0.884 & 0.284 \\
		sREFINED with MDS & 0.514 & 0.474 & 0.877 & 0.259 \\
		Xie et al. \citep{xia2018predicting}     & 1.574 & 1.295 & 0.435 & 0.991 \\
		DeepSynergy \citep{preuer2018deepsynergy}    & 1.109 & 1.058 & 0.176 & 0.929 \\
		XGBoost   \citep{friedman2002stochastic}     & 0.518 & 0.680 & 0.859 & 0.327 \\
		RF  \citep{ho1995random}           & 0.525 & 0.679 & 0.851 & 0.290 \\
		SVR \citep{drucker1997support}          & 0.561 & 0.675 & 0.830 & 0.255 \\
		EN  \citep{zou2005regularization}           & 0.618 & 0.758 & 0.789 & 0.428 \\
		\hline
	\end{tabular}%
\end{table}

Robustness analysis reveals  REFINED-CNN model stacking offers better performance as compared to single REFINED-CNN version with respect to all performance metrics. The former produced lower NRMSE between 88-90\% of times, lower NMAE between 93-95\% of times, higher PCC between 83-86\% of times, and lower Bias 78-89\% of the times. Detailed results are presented in \cref{Rob_Stack_NRMSE_ALM,Rob_Stack_NMAE_ALM,Rob_Stack_PCC_ALM,Rob_Stack_Bias_ALM} of the supplementary information. Integrated REFINED also outperformed the single-REFINED variants in considerable proportion of times. The former lowered NRMSE between 53-68\% of times, NMAE between 52-69\% of times and Bias between 43-77\% of the times, while increased PCC between 45-62\% of times.  The average results of the robustness analysis for each metric of the integrated REFINED are provided in  \cref{Rob_AM_NRMSE_ALM,Rob_AM_NMAE_ALM,Rob_AM_PCC_ALM,Rob_AM_Bias_ALM,Rob_GM_NRMSE_ALM,Rob_GM_NMAE_ALM,Rob_GM_PCC_ALM,Rob_GM_Bias_ALM} of the supplementary information. 


Gap statistics results are provided in \cref{tab:GapStatALM_NMRSE,tab:GapStatALM_NMAE,tab:GapStatALM_PCC,tab:GapStatALM_Bias} of the supplementary information. These results follow the trend observed in the NCI60 datasets with REFINED-CNN model stacking and integrated REFINED CNNs performing considerably better as compared to the single REFINED variants. The Gap statistics distribution plots per NMRSE and NMAE metrics of each model paired with the null model along with their corresponding cluster centroids for three randomly selected cell lines are provided in \cref{GapDist_NRMSE_ALM1,GapDist_NMAE_ALM1,GapDist_PCC_ALM1,GapDist_Bias_ALM1,GapDist_NRMSE_ALM2,GapDist_NMAE_ALM2,GapDist_PCC_ALM2,GapDist_Bias_ALM2,GapDist_NRMSE_ALM3,GapDist_NMAE_ALM3,GapDist_PCC_ALM3,GapDist_Bias_ALM3} of the supplementary information.

We further compare the performance of our proposed approaches with state-of-the-art models including: DeepSyenrgy \citep{preuer2018deepsynergy}, \citep{xia2018predicting}, Gradient Boosting Machine \citep{friedman2002stochastic}, Random Forests \citep{ho1995random}, Support Vector Regressor \citep{drucker1997support}, and Elastic Nets \citep{zou2005regularization}. The average performance of the models on NCI-ALMANAC dataset are provided in Table \ref{ALM_SOTA}. The detailed results including performance of each model for each cell line is provided in \cref{ALMSOTADet} of the supplementary information. Once again we observe the REFINED variants are outperforming other competing non-REFINED models for most cell lines.

\section{Discussion}
Based off \citep{kondratyuk2020ensembling,matlock2018investigation}, this study developed different ensemble learning methods for REFINED-CNN predictive methodology. Our results show that standard linear stacking of multiple single REFINED-CNN improves the prediction performance as compared to the best single REFINED CNN model. To reduce the computational cost associated with linear stacking of multiple REFINED-CNN without significantly impacting its predictive accuracy, we proposed the integrated REFINED technique.
Since each projection scheme captures a different embedded pattern of the data, the ensembling approach, associated with the integreated REFINED technique, provides a mathematical way to connect these patterns to reveal a more holistic picture. Robustness is achieved in the sense that model performance is no longer crucially dependent on the a-priori choice of the distance metric or the initial projection scheme. Furthermore, this technique offers a way to combine metric and non-metric initial dissimilarity measures via a suitable specification of the probability model for the observed distances. The integrated REFINED also offers an heuristic advantage because we can choose the probability models for observed distances by empirically observing the observed distance histograms. Different probability models for different distance metrics could be combined by the iREFINED technique to obtain the appropriate distance averaging scheme. We proved here that weighted arithmetic and geometric means turn out to be appropriate averaging schemes for common choices of distribution of observed distances.     


Through the application on both NCI60 and NCI-ALMANAC datasets, we have established the superior performance of the ensembling techniques. We benchmarked the performance of integrated REFINED with REFINED-CNN model stacking to reveal that the former produces comparable results in predicting drug sensitivity summary metrics (for example, NLOGGI50 and ComboScore) at a fraction of computation cost associated with the latter. Table \ref{time_comp} reveals computational time for REFINED-CNN model stacking is almost 4 times more than that for the integrated REFINED approach. We also observed that the integrated REFINED performed uniformly better than the REFINED image stacking model indicating the need to fix the location of feature in the set of REFINED images obtained via different projection schemes. We also proved that integrated REFINED emerges as a result of the constraint that requires the location of features, in the project 2D plane, must remain invariant under different projection schemes thereby offering an intuitive interpretation of the iREFINED technique.

\begin{table}[]
	\centering
	\caption{Execution time comparison. Comparing execution time of each step of integrated REFINED CNN model and REFINED-CNN model stacking trained on HCC\_2998 cell line data of NCI60 dataset.}
	\label{time_comp}
	\begin{tabular}{ccc}
		\hline
		Steps          & iREFINED-CNN & REFINED-CNN model stacking \\
		\hline
		MDS            &     7s     &      7s     \\
		Isomap         &     21s    &      21s    \\
		LE             &     23s    &      23s    \\
		LLE            &     28s    &      28s    \\
		NMDS + DA      &     47s    &      --     \\
		Hill climbing  & 8m \& 23s  &  33m \& 32s  \\
		CNN            & 2h \& 17m \& 36s   &   9h \& 10m \& 24s  \\
		LR             &     --     &      1s     \\
		\hline
		Total          &    2h \& 28m \& 25s   &    9h \& 45m \& 19s    \\
		\hline
	\end{tabular}
\end{table}

Of course, neither the REFINED-CNN model stacking nor integrated REFINED-CNN guarantees better performance as compared to the best single REFINED-CNN in each instance. An intuitive way to decide whether these ensembling approaches should be deployed is to assess the amount of distortion induced by each individual REFINED scheme. If it appears that particular projection is producing significantly lower distortion, we recommend fitting a single REFINED-CNN associated with that projection scheme. If, on the other hand, the distortions are similar, ensembling is advised. A formal investigation into this conjecture is a future avenue for research.

Our REFINED based architecture has two major pharmacologic implications. First, if we wish to predict the efficacy of a drug on a tumor, a rich class of regressors will consist of both numerical and image variables. The set of numeric regressors consists of (a) the chemical descriptors of the drugs, and (b) molecular characteristics of tumor that offer a genome wide profile of the tumor. The image regressors consist of histopathology images that capture the inherent heterogeneity of tumors. The REFINED technique offers a solution to this regression-on-multi-type data problem. Our technique converts all non-image regressors to legitimate images which are then processed through CNN algorithm to generate prediction. The integrated REFINED-CNN technique, developed herein, indicates that the prediction can be made robust by combining different distance metrics. Consequently, as multi-modal data collection protocols become more prevalent in the realm of pharmacogenomics, the general REFINED technique (particularly iREFINED) offers a methodology where fairly standard image based deep learning techniques can be utilized to analyze such multi-type data.

Second, we observe a high accuracy out-of-sample prediction performance of our model in NCI-ALMANAC data. This empirical predictive reliability indicates that integrated REFINED-CNN can be utilized to optimize the efficacy of a drug combination treatment regime. More specifically, given an initial choice of drug, say $D_{init}$, this technique can identify a set of drugs, from a given list of drugs, that are synergistic to $D_{init}$ in the following sense. We can keep $D_{init}$ fixed at one of the arms of the network and allow the other arm scan through the foregoing list of drugs to predict ``ComboScores''. Our bootstrapped-based inferential methodology, that enabled us to generate the intervals for NRMSE, NMAE, PCC and bias (see \cref{ALM_NRMSE_Conf,ALM_NMAE_Conf,ALM_PCC_Conf,ALM_Bias_Conf} of the supplementary information), can then be utlized to generate confidence intervals about the predicted ComboScores. This procedure can thus identify whether there exists a drug (in the list) that can be paired with $D_{init}$ to achieve significant increase in efficacy. An exploration into this line of investigation will be conducted in future.
\section*{Funding}

Research reported in this publication was supported in part by the National Institute Of General Medical Sciences of the National Institutes of Health under Award Number R01GM122084 and by the National Science Foundation under grant number CCF 2007903. The content is solely the responsibility of the authors and does not necessarily represent the official views of the National Institutes of Health or National Science Foundation.\vspace*{-12pt} \\

\textit{Conflict of Interest:} none declared.

\bibliographystyle{natbib}

\bibliography{main}

\end{document}